\documentclass[11pt]{article}

\usepackage[preprint]{acl}

\usepackage{times}
\usepackage{latexsym}

\usepackage[T1]{fontenc}

\usepackage[utf8]{inputenc}

\usepackage{microtype}

\usepackage{inconsolata}

\usepackage{url}
\usepackage{graphicx}
\usepackage{booktabs}
\usepackage{multirow}
\usepackage{tikz}
\usepackage{pgfplots}
\usepackage{xcolor}
\usepackage{amsmath}
\usepackage{amssymb}
\usepackage{tikz}
\usepackage{xcolor}
\usepackage[table,xcdraw]{xcolor} 
\usepackage{amsmath}
\usepackage{marvosym,fontawesome}
\usepackage{pgfplots}
\usepgfplotslibrary{groupplots}
\usetikzlibrary{positioning, shapes, arrows, patterns}
\usepackage{subcaption}

\definecolor{color0}{RGB}{76,175,80}   
\definecolor{color1}{RGB}{255,193,7}   
\definecolor{color2}{RGB}{244,67,54}   

\newcommand{\dataset}{Atomic-SNLI}
\newcommand{\snli}{SNLI}

\tikzset{
  mynode/.style={rounded corners=4pt, align=center, text width=\boxwidth,
                 minimum height=\boxheight, font=\large},
  title/.style={font=\Large\bfseries, text=black!80},
  box/.style={draw=black!30, rounded corners=6pt, inner sep=6pt,
              fill=white, drop shadow={shadow xshift=1pt, shadow yshift=-1pt,
              fill=black!15}}
}

\newcommand{\materialCards}[3][\textwidth]{%
  \centering
  \begin{tikzpicture}[rounded corners=2pt]
    \node[
      draw      = none,
      fill      = none,
      text      = black,
      text width=\textwidth,
      minimum height = 1.1cm,
      align     = center,
      font      = \bfseries\sffamily
    ] (title) {#2};

    \node[
      draw      = none,
      fill      = yellow!5,
      text      = black,
      text width=\textwidth,
      align     = left,
      anchor    = north west,
      font      = \normalsize
    ] (desc) at (title.south west) {#3};

    \draw[black!60, line width=1pt] ([xshift=2pt]desc.north west) -- ([xshift=-2pt]desc.north east);

    \draw[black!60, line width=1pt] ([xshift=2pt]title.north west) -- ([xshift=-2pt]title.north east);

    \draw[black!60, line width=1pt] ([xshift=2pt]desc.south west) -- ([xshift=-2pt]desc.south east);
  \end{tikzpicture}%
}

\title{Atomic-SNLI: Fine-Grained Natural Language Inference through Atomic Fact Decomposition}

\author{Minghui Huang \\
  The University of Texas at Austin  \\
  \texttt{minghuihuang@utexas.edu} \\}

\begin{document}

\maketitle

\begin{abstract}
Current Natural Language Inference (NLI) systems primarily operate at the sentence level, providing black-box decisions that lack explanatory power. While atomic-level NLI offers a promising alternative by decomposing hypotheses into individual facts, we demonstrate that the conventional assumption that a hypothesis is entailed only when all its atomic facts are entailed fails in practice due to models' poor performance on fine-grained reasoning. Our analysis reveals that existing models perform substantially worse on atomic level inference compared to sentence level tasks. To address this limitation, we introduce Atomic-SNLI, a novel dataset constructed by decomposing SNLI and enriching it with carefully curated atomic level examples through linguistically informed generation strategies. Experimental results demonstrate that models fine-tuned on Atomic-SNLI achieve significant improvements in atomic reasoning capabilities while maintaining strong sentence level performance, enabling both accurate judgements and transparent, explainable results at the fact level.\footnote{Data available at: \url{https://huggingface.co/datasets/MinghuiHuang/AtomicSnli}}
\end{abstract}

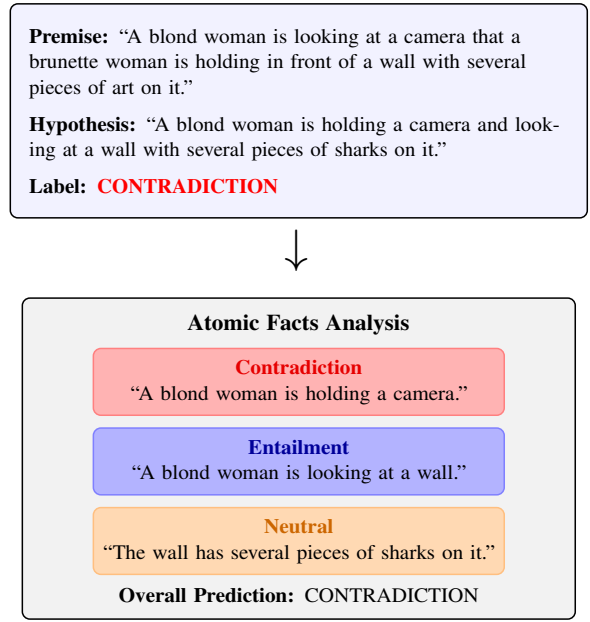
\begin{figure}[t]
\centering
\resizebox{0.48\textwidth}{!}{%
\begin{tikzpicture}[
    box/.style={draw, thick, rounded corners, fill=white, align=left},
    factbox/.style={
        draw=#1!50, thick, fill=#1!30, rounded corners,
        inner sep=6pt, minimum height=1cm, text width=7.2cm, align=center
    },
    section/.style={font=\large\bfseries, align=center}
]

\node[box, fill=blue!5, inner sep=10pt, text width=10cm, minimum height=4cm] (premise) at (0,2) {
    \textbf{Premise:} ``A blond woman is looking at a camera that a brunette woman is holding in front of a wall with several pieces of art on it.''
    
    \medskip
    \textbf{Hypothesis:} ``A blond woman is holding a camera and looking at a wall with several pieces of sharks on it.''
    
    \medskip
    \textbf{Label:} \textcolor{red}{\bfseries CONTRADICTION}
};

\node[align=center] (label) at (0,-0.6) {
    \textcolor{black}{\Huge \textbf{$\downarrow$}}
};
\node[box, fill=gray!10, text width=10cm, minimum height=6cm, align=center] (facts) at (0,-4.5) {
    \textbf{\large Atomic Facts Analysis}
    
    \medskip
    \begin{tabular}{c}
      \medskip
        \tikz[baseline]{\node[factbox=red] {
            \textcolor{red!90!black}{\bfseries Contradiction}\\
            ``A blond woman is holding a camera.''
        };} \\[1em]
        \medskip
        \tikz[baseline]{\node[factbox=blue] {
            \textcolor{blue!60!black}{\bfseries Entailment}\\
            ``A blond woman is looking at a wall.''
        };} \\[1em]
        \tikz[baseline]{\node[factbox=orange] {
            \textcolor{orange!80!black}{\bfseries Neutral}\\
            ``The wall has several pieces of sharks on it.''
        };}
    \end{tabular}
    
    \medskip
    \textbf{Overall Prediction:} CONTRADICTION
};

\end{tikzpicture}
}
\caption{
    How Sentence-Level and Atomic-Level NLI Explains Reasoning. \\ \textbf{Top (Sentence-Level):} The model reads the full premise and hypothesis and gives one answer: "contradiction." The reason for this decision is unclear. \\ \textbf{Bottom (Atomic-Level):} The hypothesis is broken into single facts. The model checks each fact against the premise. These individual checks are combined to reach the final "contradiction" verdict, clearly showing which part caused the conflict.
}
\label{fig:nli_analysis}
\end{figure}

\section{Introduction}
\label{sec:intro}

Natural Language Inference (NLI) has emerged as a core task for evaluating semantic understanding in NLP, requiring a system to determine the logical relationship (entailment, contradiction, or neutrality) between a premise and a hypothesis \cite{bowman_large_2015,williams_broad-coverage_2018}. However, traditional NLI operates at the sentence level, presenting a critical limitation in interpretability: while a model can predict a label for a premise-hypothesis pair, it is often impossible to discern which specific parts of the hypothesis led to the overall decisions. This black-box nature hinders trust and diagnostic analysis.

\textbf{Atomic-level NLI} addresses this by decomposing complex reasoning into fine-grained steps. This approach breaks down a hypothesis $h$ into a set of minimal, self-contained atomic facts $\mathcal{A}_h$ and judges the relationship between the premise $p$ and each individual fact $a_i$. The final sentence-level label is then derived through a bottom-up, compositional process by aggregating these atomic predictions using a logical inference rule. This framework promises transparent and explainable decisions, as illustrated in Figure~\ref{fig:nli_analysis}.

A widely adopted principle in fact verification and NLI \cite{huang_decmetrics:_2025,tang_minicheck:_2024} formalizes this intuition with the following rule:
\begin{quote}
\itshape
A hypothesis $h$ is \textbf{supported} by $p$ if and only if \textbf{every} atomic fact $a \in \mathcal{A}_h$ is supported by $p$; otherwise $h$ is \textbf{not supported}.
\end{quote}

Surprisingly, our analysis reveals a fundamental disconnect between this theoretical ideal and the practical capabilities of current models. We find that the atomic-level accuracy of state-of-the-art NLI models is comparable to, or even lower than, their sentence-level accuracy. This indicates that even if a model were to achieve perfect atomic-level judgments, the standard inference rule would often fail to produce the correct sentence-level label, highlighting a critical weakness in compositional reasoning.

This work makes three key contributions:
\begin{enumerate}
\item We systematically investigate the limitations of compositional reasoning in current NLI models through a fine-grained, atomic-level analysis.
\item We introduce \dataset, a large-scale dataset for atomic-level NLI, constructed by decomposing \snli~and enriching it with high-quality, automatically generated neutral and contradictory atomic examples.
\item We demonstrate that fine-tuning standard NLI models on \dataset~significantly enhances their atomic reasoning capabilities while maintaining strong performance on the standard sentence-level NLI task, effectively bridging the gap between fine-grained and holistic understanding.
\end{enumerate}

As illustrated in Figure~\ref{fig:nli_analysis}, atomic-level analysis reveals that the overall contradiction judgment stems from a single critical conflict ("A blond woman is holding a camera"), while other components are correctly identified as entailed or neutral. This granular perspective provides a transparent and interpretable window into the model's reasoning process.

\section{Related Work}
\label{sec:related}

\paragraph{Fact Decomposition in NLP.} Decomposing text into atomic facts has proven valuable across various NLP applications. \citet{min_factscore:_2023} employ fact decomposition for factuality evaluation, while \citet{tang_minicheck:_2024} and \citet{huang_decmetrics:_2025} leverage it in automated fact-checking pipelines. \citet{chen_dense_2024} demonstrate that using decomposed atomic facts as retrieval units improves performance in dense retrieval and downstream QA systems. Most relevant to our work, \citet{srikanth_nli_2025} explore hypothesis decomposition in NLI, finding that LLMs struggle with logical consistency on atomic NLI tasks.

\paragraph{Decomposition Methods.} Most existing approaches employ LLMs to generate atomic facts from input text \cite{min_factscore:_2023,kamoi_wice:_2023,tang_minicheck:_2024,srikanth_nli_2025}. However, \citet{huang_decmetrics:_2025} argue that LLM-generated decompositions lack stability and reliability. They propose \textsc{DecMetrics} to evaluate fact decomposition quality and use it as a reward signal to train a specialized \textsc{DecModel} for text decomposition.

\paragraph{Fine-grained NLI.} While standard NLI operates at sentence level, recent work has explored more granular reasoning~\cite{stacey_atomic_2024,srikanth_nli_2025}. Our approach differs by focusing specifically on atomic fact-level entailment and providing a dedicated dataset and evaluation framework for this task.

\section{Atomic-Level NLI Analysis}
\label{sec:atomic}

Traditional NLI datasets contain instances $(p, h, y)$, where $p$ is a premise, $h$ is a hypothesis, and $y \in \{\text{entailment, contradiction, neutral}\}$ is the entailment label. In \snli, hypotheses are formulated at sentence level. To enable atomic-level prediction, we first decompose each hypothesis into atomic facts, then aggregate predictions across these atomic components.

\subsection{Hypothesis Decomposition}

Using \textsc{DecModel}~\cite{huang_decmetrics:_2025}, we decompose each hypothesis $h$ in the \snli~test set into a set of atomic facts $\mathcal{A}_h = \{a_1, a_2, \dots, a_n\}$. After filtering out invalid examples, we obtain 9,824 valid hypotheses for analysis.

Table~\ref{tab:atomic_stats} presents the distribution of atomic facts per hypothesis. The decomposition yields between 1 and 5 atomic facts per hypothesis, with a strong concentration at the lower end: 8,767 hypotheses (89.2\%) contain only a single atomic fact, while only 27 hypotheses (0.3\%) contain 4 or more atomic facts.

\begin{table}[t]
\centering
\small
\begin{tabular}{cc}
\toprule
\textbf{\# Atoms} & \textbf{\# Instances} \\
\midrule
1 & 8,767 \\
2 & 881 \\
3 & 149 \\
4 & 25 \\
5 & 2 \\
\midrule
\textbf{Total} & \textbf{9,824} \\
\bottomrule
\end{tabular}
\caption{Distribution of atomic facts per hypothesis in the decomposed \snli~test set. The majority of hypotheses (89.2\%) consist of only one atomic fact.}
\label{tab:atomic_stats}
\end{table}

\begin{table}[t]
\centering
\small
\begin{tabular}{lcc}
\toprule
\textbf{Model} & \textbf{Sentence-Level} & \textbf{Atomic-Level} \\
\midrule
nli-deberta-v3-base & 92.38 & 91.65 \\
nli-deberta-v3-small & 91.65 & 91.04 \\
nli-deberta-v3-xsmall & 91.64 & 90.95 \\
nli-MiniLM2-L6-H768 & 91.37 & 90.58 \\
\bottomrule
\end{tabular}
\caption{Comparison of sentence-level and atomic-level NLI accuracy (\%). Atomic-level performance consistently lags behind sentence-level performance across all models.}
\label{tab:nli_comparison}
\end{table}

\begin{figure*}[t]
\centering
\begin{tikzpicture}
    \begin{groupplot}[
        group style={
            group size=1 by 3,
            vertical sep=1.22cm,
            ylabels at=edge left,
            x descriptions at=edge bottom
        },
        ybar=0.2pt,
        width=0.8\textwidth,
        height=0.21\textwidth,
        xlabel={Number of Atomic Facts},
        ylabel={Percentage (\%)},
        ymin=40, ymax=105,
        ytick={0,20,40,60,80,100},
        xtick={1,2,3,4,5},
        enlarge x limits=0.15,
        ymajorgrids=true,
        grid style={dashed, gray!30},
        tick label style={font=\footnotesize},
        label style={font=\small},
        title style={font=\small\bfseries, yshift=5pt},
        legend style={font=\tiny}
    ]

    \nextgroupplot[title={Entailment}]
        \addplot[fill=blue!50, draw=blue!50] 
            coordinates {(1,94.91) (2,92.40) (3,90.60) (4,76.00) (5,100.00)};
        \addplot[fill=red!50, draw=red!50] 
            coordinates {(1,92.87) (2,94.81) (3,94.12) (4,100.00) (5,100.00)};
        \addplot[fill=orange!50, draw=orange!50] 
            coordinates {(1,92.57) (2,78.21) (3,55.17) (4,50.00) (5,100.00)};

    \nextgroupplot[title={Neutral}]
        \addplot[fill=blue!50, draw=blue!50] 
            coordinates {(1,93.02) (2,87.40) (3,87.25) (4,76.00) (5,100.00)};
        \addplot[fill=red!50, draw=red!50] 
            coordinates {(1,88.71) (2,85.68) (3,91.30) (4,64.29) (5,100.00)};
        \addplot[fill=orange!50, draw=orange!50] 
            coordinates {(1,89.16) (2,84.53) (3,82.89) (4,90.00) (5,0.00)};

    \nextgroupplot[title={Contradiction}]
        \addplot[fill=blue!50, draw=blue!50] 
            coordinates {(1,96.90) (2,92.51) (3,87.25) (4,92.00) (5,100.00)};
        \addplot[fill=red!50, draw=red!50] 
            coordinates {(1,95.44) (2,80.60) (3,69.84) (4,60.00) (5,100.00)};
        \addplot[fill=orange!50, draw=orange!50] 
            coordinates {(1,95.31) (2,96.79) (3,100.00) (4,100.00) (5,100.00)};

    \end{groupplot}
\end{tikzpicture}
\resizebox{0.45\textwidth}{!}{
\begin{tabular}{cccccc}
    \tikz[baseline]{\draw[fill=blue!50, draw=blue!50] (0,0) rectangle (0.4,0.4);} & Accuracy &
    \tikz[baseline]{\draw[fill=red!50, draw=red!50] (0,0) rectangle (0.4,0.4);} & Precision &
    \tikz[baseline]{\draw[fill=orange!50, draw=orange!50] (0,0) rectangle (0.4,0.4);} & Recall \\
\end{tabular}
}
\caption{Performance metrics (Accuracy, Precision, Recall) of nli-deberta-v3-base across different numbers of atomic facts per hypothesis, categorized by NLI labels.}
\label{fig:atomic_performance_subplots}
\end{figure*}
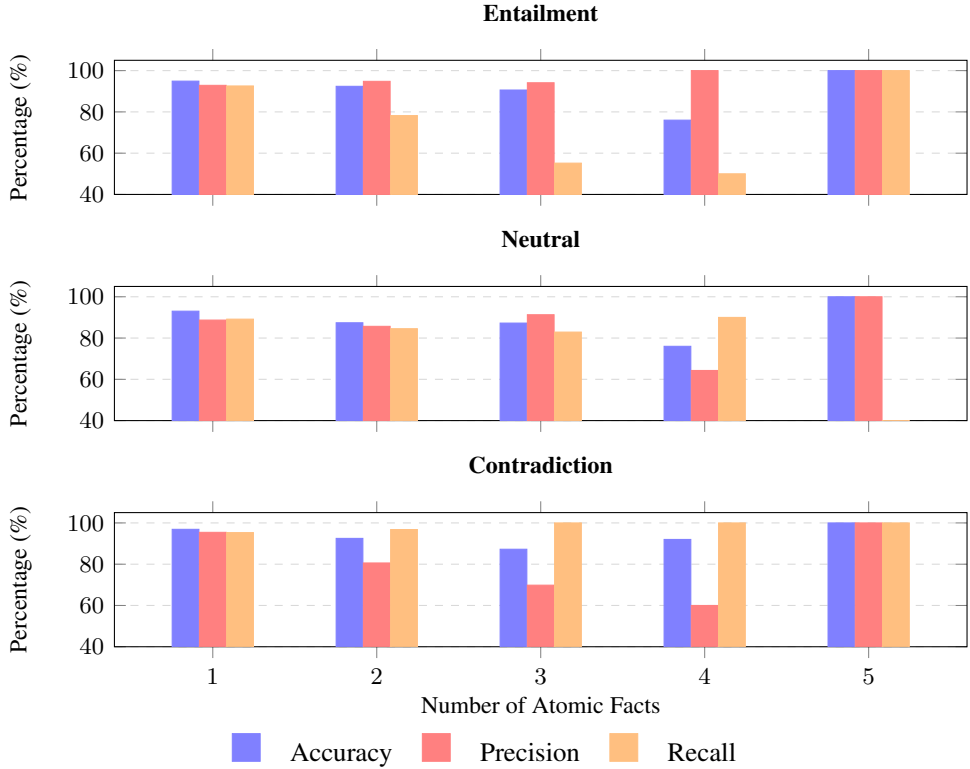
\subsection{Inference Framework}

By decomposing each hypothesis $h$ into a set $\mathcal{A}_h$ of minimal atomic facts, we can precisely identify which components are supported or refuted by premise $p$. This decomposition enables transparent and interpretable predictions of the overall entailment relationship.

\paragraph{Inference Rules.}
For each instance $(p, h, \mathcal{A}_h)$, we define the following compositional inference rules:
\begin{enumerate}
    \item \textbf{Entailment:} $h$ is entailed by $p$ if and only if \textit{every} atomic fact $a \in \mathcal{A}_h$ is entailed by $p$.
    \item \textbf{Contradiction:} $h$ contradicts $p$ if \textit{at least one} atomic fact $a \in \mathcal{A}_h$ contradicts $p$.
    \item \textbf{Neutral:} $h$ is neutral with respect to $p$ if at least one atomic fact is neutral toward $p$, and \textit{no} atomic fact contradicts $p$.
\end{enumerate}

\subsection{Experimental Setup}

We evaluate four pre-trained NLI models from the Sentence Transformers~\cite{reimers_sentence-bert:_2019} cross-encoder collection:\footnote{\url{https://www.sbert.net/docs/pretrained_cross-encoders.html}}

\begin{itemize}
    \item \textbf{nli-deberta-v3-base}: Base-sized DeBERTa-V3 trained on SNLI~\cite{bowman_large_2015} and MultiNLI~\cite{williams_broad-coverage_2018}
    \item \textbf{nli-deberta-v3-small}: Small-sized DeBERTa-V3 variant with identical training
    \item \textbf{nli-deberta-v3-xsmall}: Extra-small-sized DeBERTa-V3 with identical training  
    \item \textbf{nli-MiniLM2-L6-H768}: MiniLM-L6 architecture with 6 layers and 768 hidden dimensions, distilled on SNLI and MultiNLI datasets
\end{itemize}

We compare sentence-level NLI performance against atomic-level NLI performance on the \snli~dataset.

\subsection{Results and Analysis}

As shown in Table~\ref{tab:nli_comparison}, atomic-level NLI achieves lower overall performance compared to sentence-level NLI across all models, with performance gaps ranging from 0.61 to 0.79 percentage points.

Further analysis with the nli-deberta-v3-base model (Figure~\ref{fig:atomic_performance_subplots}) reveals performance variation across different numbers of atomic facts. For multi-fact hypotheses, we observe increased recall for contradiction and increased precision for entailment, aligning with our inference rules which make atomic-level predictions more conservative for entailment.

\paragraph{Model Biases and Limitations.} Our analysis reveals that existing NLI models often fail to adhere to the strict logical constraints of fine-grained inference. We hypothesize this stems from misalignment between training objectives and compositional reasoning requirements. Models trained on standard benchmarks learn dataset-specific heuristics and tend to over-predict contradiction when identifying any contradictory evidence. This bias results in artificially high recall for contradiction under deterministic inference rules.

\section{Atomic-SNLI Dataset Construction}
\label{sec:asnli}

To address the limitations of coarse-grained sentence-level inference and enable fine-grained model training, we introduce \textbf{Atomic-SNLI}, a large-scale dataset for atomic-level natural language inference. We construct this resource by systematically decomposing the SNLI dataset~\cite{bowman_large_2015} to generate triplets of the form $(p, a, y_{\text{atomic}})$, where a premise $p$ is paired with an atomic fact $a$ and a fine-grained label $y_{\text{atomic}} \in {\text{entailment, contradiction, neutral}}$. This structure explicitly trains models to perform precise, localized reasoning over individual factual claims, isolating them from the compositional complexity of full sentences.

Our construction process is applied to the training and validation splits of SNLI to create supervised training data for atomic reasoning. Crucially, to evaluate a model's ability to compose atomic inferences into a sentence-level judgment, the test set is structured differently. It retains the original format $(p, \mathcal{A}_h, y)$, where $\mathcal{A}_h$ is the set of atomic facts decomposed from the hypothesis $h$. The final entailment judgment $\hat{y}$ is then predicted by aggregating the atomic predictions for all $a_i \in \mathcal{A}_h$ using the inference rules defined in Section~\ref{sec:atomic}.

\subsection{Data Construction Methodology}
\begin{figure}[t]
\centering
\resizebox{0.5\textwidth}{!}{%
\begin{tikzpicture}[
    node distance=0.8cm,
    box/.style={draw, thick, rounded corners=4pt, align=center},
    process/.style={box, minimum width=3cm, minimum height=1.5cm},
    data/.style={box, minimum width=3cm, minimum height=1.2cm},
    special/.style={box, minimum width=5cm, minimum height=1.4cm},
    label/.style={font=\small\bfseries, align=center},
    arrow/.style={->, >=stealth, thick, line width=1.5pt},
    branch/.style={font=\large\bfseries, text width=2.5cm, align=center}
]

\draw[thick, blue!50, rounded corners=8pt,fill=gray!10] (-6,-4) rectangle (-2.2,-10.5);
\draw[thick, orange!50, rounded corners=8pt,fill=gray!10] (-2,-4) rectangle (1.9,-10.5);
\draw[thick, red!50, rounded corners=8pt,fill=gray!10] (2.1,-4) rectangle (6.2,-10.5);

\node[data, fill=purple!8, minimum width=8cm] (original) at (0,2) 
    {\textbf{SNLI Dataset} $(p, h)$};

\node[data, fill=purple!10, minimum width=8cm] (decompose) at (0,0.5) 
    {\textbf{Decomposition} with \textsc{DecModel}};

\node[data, fill=purple!8, minimum width=8cm] (atoms) at (0,-1) 
    {\textbf{Atomic Facts} $\mathcal{A}_h$};

\draw[arrow, black] (original) -- (decompose);
\draw[arrow, black] (decompose) -- (atoms);

\draw[thick, purple!40, dashed, rounded corners=8pt] (6.5,4) rectangle (-6.5,-2);
\node[align=center, purple!70!black,font=\Large] at (0,3.5) {\textcolor{purple}{\Large \faStarO} \textbf{Step 1: Fact Decomposition}};

\node[branch, text=blue!80!black] (entail-label) at (-4.2,-4.5) {DIRECT};
\node[process, fill=blue!18] (entail-process) at (-4,-6) 
    {\textbf{Automatic Pairing}};
\node[process, fill=blue!18] (entail-filter) at (-4,-7.8) 
    {\textbf{Filter} \\ $\tau_{rel} > 0.5$ };
\node[data, fill=blue!12] (entail-data) at (-4,-9.5) 
    {\textbf{Direct Pairs} \\ $(p, a)$};
\draw[arrow, blue!80!black] (entail-process) -- (entail-filter);
\draw[arrow, blue!80!black] (entail-filter) -- (entail-data);

\node[branch, text=orange!80!black] (neutral-label) at (0,-4.5) {NEUTRAL};
\node[process, fill=orange!18] (neutral-retrieve) at (0,-6) 
    {\textbf{Retrieval}};
\node[process, fill=orange!15] (neutral-filter) at (0,-7.8) 
    {\textbf{Filter} \\ $\tau_n > 0.5$};
\node[data, fill=orange!12] (neutral-data) at (0,-9.5) 
    {\textbf{Neutral Pairs} \\ $(p, a'')$};
\draw[arrow, orange!80!black] (neutral-retrieve) -- (neutral-filter);
\draw[arrow, orange!80!black] (neutral-filter) -- (neutral-data);

\node[branch, text=red!80!black] (contradict-label) at (3.5,-4.5) {CONTRADICTION};
\node[process, fill=red!18] (contradict-gen) at (4,-6) 
    {\textbf{LLM Generation}};
\node[process, fill=red!15] (contradict-filter) at (4,-7.8) 
    {\textbf{Filter} \\ $\tau_c > 0.5$};
\node[data, fill=red!12] (contradict-data) at (4,-9.5) 
    {\textbf{Contradiction Pairs} \\ $(p, a')$};

\draw[arrow, red!80!black] (contradict-gen) -- (contradict-filter);
\draw[arrow, red!80!black] (contradict-filter) -- (contradict-data);

\draw[thick, gray!50, dashed, rounded corners=8pt] (-6.5,-2.5) rectangle (6.5,-11);
\node[align=center, blue!70!black,font=\Large] at (0,-3) {\textcolor{blue}{\Large \faStarHalfO} \textbf{Step 2: Label-specific Processing}};

\draw[thick, yellow!70!black, dashed, rounded corners=8pt] (-6.5,-11.5) rectangle (6.5,-14.5);
\node[special, fill=yellow!30, minimum width=10cm] (final) at (0,-13.5) 
    {\textbf{Atomic-SNLI Dataset} $(p, a, y_{atomic})$};
\node[align=center, yellow!70!black,font=\Large] at (0,-12) {\textcolor{yellow!70!black}{\Large \faStar} \textbf{Step 3: Dataset Assembly}};

\end{tikzpicture}
}
\caption{\dataset~construction pipeline: Original \snli~pairs are decomposed into atomic facts, then processed through three label-specific pipelines to generate high-quality entailment, contradiction, and neutral pairs. $\tau_{rel} \in \{\tau_e, \tau_n, \tau_c\}$.}
\label{fig:data_generation}
\end{figure}
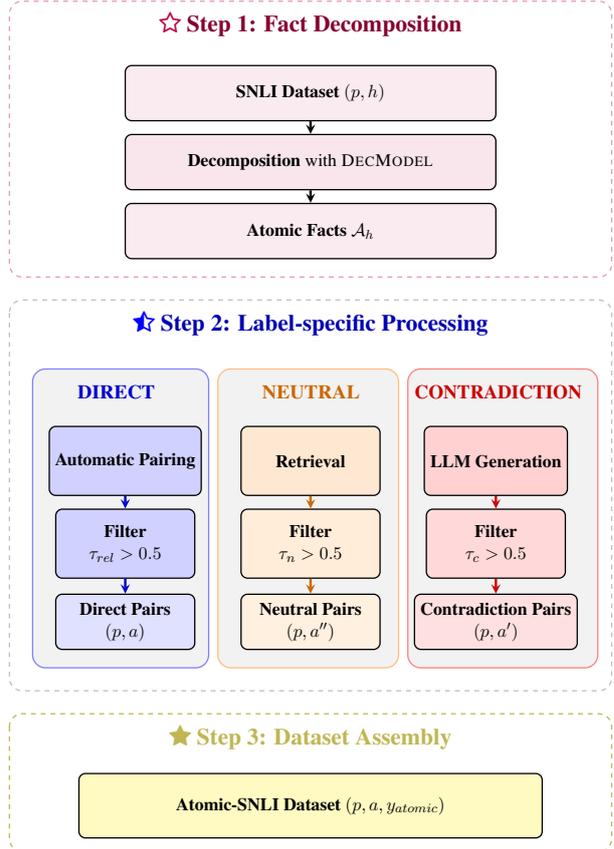

The Atomic-SNLI dataset is constructed from SNLI premise-hypothesis pairs $(p, h)$ through a multi-stage pipeline that ensures data quality and coverage across all three NLI relations.

\paragraph{Hypothesis Decomposition.} Each hypothesis $h$ in original SNLI is decomposed into atomic facts $\mathcal{A}_h = \{a_1, a_2, \dots, a_n\}$ using \textsc{DecModel}~\cite{huang_decmetrics:_2025}. Each atomic fact $a_i$ represents a single, self-contained proposition that contributes to the overall meaning of $h$. Single-fact hypotheses are directly converted into atomic triplets $(p, a, y)$, while multi-fact hypotheses undergo specialized processing for each label type.

\paragraph{Entailment Pairs Construction}
For SNLI pairs labeled \textbf{entailment}, we automatically assign the entailment label to all corresponding decomposed atomic facts $(p, a)$ where $a \in \mathcal{A}_h$. We apply quality filtering using an existing NLI model (nli-deberta-v3-base) to retain only pairs with entailment confidence $\tau_e > 0.5$. This provides high-quality positive examples, as the premise $p$ must support each atomic component of $h$ by definition of the entailment relationship.

\paragraph{Neutral Pairs Construction}
We employ a hybrid approach combining direct filtering and retrieval-based generation:
\begin{itemize}
    \item \textbf{Direct Filtering:} For existing $(p,a)$ pairs, we use nli-deberta-v3-base to compute confidence scores $\{\tau_e, \tau_n, \tau_c\}$ and retain pairs with neutrality confidence $\tau_n > 0.5$
    \item \textbf{Retrieval-based Generation:} To expand neutral examples, we implement:
    \begin{enumerate}
        \item \textbf{Candidate Retrieval:} Using BM25~\cite{trotman_improvements_2014}, retrieve top-100 lexically similar atomic facts from other instances based on premise $p$
        \item \textbf{Semantic Re-ranking:} A cross-encoder model~\cite{reimers_sentence-bert:_2019} re-ranks candidates by neutrality score $\tau_n$
        \item \textbf{Quality Validation:} Filter pairs where $\tau_n > 0.5$ to ensure genuine neutrality
    \end{enumerate}
\end{itemize}

\paragraph{Contradiction Pairs Construction}
We combine direct extraction with LLM-based generation:
\begin{itemize}
    \item \textbf{Direct Extraction:} For existing $(p,a)$ pairs, we use nli-deberta-v3-base to identify contradictions with confidence $\tau_c > 0.5$
    \item \textbf{LLM-based Generation:} To enhance contradiction diversity:
    \begin{enumerate}
        \item \textbf{Generation:} For each premise $p$ and entailed atomic fact $a$, we prompt Qwen3-32B~\cite{yang_qwen3_2025} to generate minimally altered versions $a'$ that contradict $p$ while preserving grammaticality and lexical similarity
        \item \textbf{Validation:} Generated pairs $(p, a')$ are evaluated by an ensemble of NLI models, retaining only pairs where all models predict ``contradiction'' with confidence $\tau_c > 0.5$
    \end{enumerate}
\end{itemize}

\subsection{Dataset Statistics}
 
\begin{table*}[t]
\centering
\small
\begin{tabular}{llcccc}
\toprule
\textbf{Split} & \textbf{Class} & \textbf{\# Instances} & \textbf{\# Unique Premises} & \textbf{Hypo. Avg. Len.} & \textbf{Premise Use} \\
\midrule
\multirow{5}{*}{Train}
& Entailment & 170,917 & 141,174 & 32.09 & 1.21 \\
& Neutral & 232,076 & 149,078 & 38.73 & 1.56 \\
& Contradiction & 222,288 & 149,896 & 35.24 & 1.48 \\
\cmidrule(lr){2-6}
& \textbf{Atomic-SNLI Total} & \textbf{625,281} & \textbf{150,735} & \textbf{35.67} & \textbf{4.15} \\
& \textbf{SNLI Total} & \textbf{549,367} & \textbf{150,736} & \textbf{37.47} & \textbf{3.64} \\
\midrule
\multirow{5}{*}{Validation}
& Entailment & 3,085 & 2,877 & 32.56 & 1.07 \\
& Neutral & 4,105 & 2,965 & 38.67 & 1.38 \\
& Contradiction & 4,145 & 3,171 & 35.18 & 1.31 \\
\cmidrule(lr){2-6}
& \textbf{Atomic-SNLI Total} & \textbf{11,335} & \textbf{3,317} & \textbf{35.73} & \textbf{3.42} \\
& \textbf{SNLI Total} & \textbf{9,842} & \textbf{3,319} & \textbf{37.90} & \textbf{2.96} \\
\midrule
\multirow{2}{*}{Test}
& \textbf{Atomic-SNLI Total} & \textbf{11,086} & \textbf{3,323} & \textbf{35.47} & \textbf{3.34} \\
& \textbf{SNLI Total} & \textbf{9,824} & \textbf{3,323} & \textbf{37.70} & \textbf{2.96} \\
\bottomrule
\end{tabular}
\caption{Statistics of the Atomic-SNLI dataset. For each split, we show the distribution across relation classes (entailment, neutral, contradiction) and the overall totals. The SNLI Total row provides the corresponding statistics from the original SNLI dataset for comparison. The ``Premise Use'' column indicates the average number of hypotheses generated from a single premise.}
\label{tab:dataset_stats}
\end{table*}

The constructed Atomic-SNLI dataset offers a rich resource for fine-grained natural language inference. Key statistics are summarized in Table~\ref{tab:dataset_stats}. Across all splits—training, validation, and test—the dataset retains the same set of premises as the original SNLI corpus. However, the total number of inference instances (i.e., premise–atomic hypothesis pairs $(p, a)$) is substantially larger than the number of premise–hypothesis pairs $(p, h)$ in SNLI. This expansion arises directly from our decomposition process: each sentence-level hypothesis in SNLI is systematically broken down into multiple atomic facts, thereby increasing data granularity. 

This fine-grained representation constitutes the core contribution of Atomic-SNLI, offering models a more precise and challenging signal to learn the fundamental building blocks of textual entailment. Moreover, our balanced sampling strategy during the generation of neutral and contradictory atomic hypotheses ensures a well-distributed label space, preventing any single relation type from dominating the training data.

\section{Experiments}
\label{sec:experiments}

\begin{table*}[t]
\centering
\small
\begin{tabular}{lllllll}
\toprule
\textbf{Atomic Num} & \textbf{Model} & \textbf{Dataset} & \textbf{Accuracy} & \textbf{Precision} & \textbf{Recall} & \textbf{F1 Score} \\
\midrule

\multirow{6}{*}{2} & \multirow{2}{*}{DeBERTa-v3-xsmall}
& SNLI & 84.22 & 84.60 & 84.96 & 84.52 \\
& & \dataset & 85.93 $\color{green!80!black}{\scriptstyle\uparrow 1.71}$ & 86.18 $\color{green!80!black}{\scriptstyle\uparrow 1.58}$ & 86.23 $\color{green!80!black}{\scriptstyle\uparrow 1.27}$ & 86.19 $\color{green!80!black}{\scriptstyle\uparrow 1.67}$ \\
\cmidrule{2-7}
& \multirow{2}{*}{ELECTRA-small}
& SNLI & 81.61 & 81.77 & 82.45 & 81.86 \\
& & \dataset & 83.09 $\color{green!80!black}{\scriptstyle\uparrow 1.48}$ & 83.11 $\color{green!80!black}{\scriptstyle\uparrow 1.34}$ & 83.45 $\color{green!80!black}{\scriptstyle\uparrow 1.00}$ & 83.24 $\color{green!80!black}{\scriptstyle\uparrow 1.38}$ \\
\cmidrule{2-7}
& \multirow{2}{*}{BERT-base}
& SNLI & 84.34 & 84.33 & 85.06 & 84.53 \\
& & \dataset & 84.79 $\color{green!80!black}{\scriptstyle\uparrow 0.45}$ & 84.87 $\color{green!80!black}{\scriptstyle\uparrow 0.54}$ & 85.35 $\color{green!80!black}{\scriptstyle\uparrow 0.29}$ & 85.01 $\color{green!80!black}{\scriptstyle\uparrow 0.48}$ \\
\midrule

\multirow{6}{*}{3} & \multirow{2}{*}{DeBERTa-v3-xsmall}
& SNLI & 76.51 & 76.54 & 78.86 & 75.64 \\
& & \dataset & 86.58 $\color{green!80!black}{\scriptstyle\uparrow 10.07}$ & 84.46 $\color{green!80!black}{\scriptstyle\uparrow 7.92}$ & 83.24 $\color{green!80!black}{\scriptstyle\uparrow 4.38}$ & 83.80 $\color{green!80!black}{\scriptstyle\uparrow 8.16}$ \\
\cmidrule{2-7}
& \multirow{2}{*}{ELECTRA-small}
& SNLI & 71.81 & 71.98 & 74.84 & 71.23 \\
& & \dataset & 79.87 $\color{green!80!black}{\scriptstyle\uparrow 8.06}$ & 76.72 $\color{green!80!black}{\scriptstyle\uparrow 4.74}$ & 77.11 $\color{green!80!black}{\scriptstyle\uparrow 2.27}$ & 76.76 $\color{green!80!black}{\scriptstyle\uparrow 5.53}$ \\
\cmidrule{2-7}
& \multirow{2}{*}{BERT-base}
& SNLI & 75.17 & 74.39 & 77.35 & 74.08 \\
& & \dataset & 82.55 $\color{green!80!black}{\scriptstyle\uparrow 7.38}$ & 80.02 $\color{green!80!black}{\scriptstyle\uparrow 5.63}$ & 78.47 $\color{green!80!black}{\scriptstyle\uparrow 1.12}$ & 78.84 $\color{green!80!black}{\scriptstyle\uparrow 4.76}$ \\
\midrule

\multirow{6}{*}{4} & \multirow{2}{*}{DeBERTa-v3-xsmall}
& SNLI & 88.00 & 78.57 & 74.44 & 75.68 \\
& & \dataset & 72.00 $\color{red!80!black}{\scriptstyle\downarrow 16.00}$ & 62.63 $\color{red!80!black}{\scriptstyle\downarrow 15.94}$ & 62.78 $\color{red!80!black}{\scriptstyle\downarrow 11.66}$ & 62.59 $\color{red!80!black}{\scriptstyle\downarrow 13.09}$ \\
\cmidrule{2-7}
& \multirow{2}{*}{ELECTRA-small}
& SNLI & 88.00 & 78.57 & 74.44 & 75.68 \\
& & \dataset & 76.00 $\color{red!80!black}{\scriptstyle\downarrow 12.00}$ & 83.81 $\color{green!80!black}{\scriptstyle\uparrow 5.24}$ & 66.67 $\color{red!80!black}{\scriptstyle\downarrow 7.77}$ & 68.69 $\color{red!80!black}{\scriptstyle\downarrow 6.99}$ \\
\cmidrule{2-7}
& \multirow{2}{*}{BERT-base}
& SNLI & 88.00 & 78.57 & 74.44 & 75.68 \\
& & \dataset & 64.00 $\color{red!80!black}{\scriptstyle\downarrow 24.00}$ & 56.57 $\color{red!80!black}{\scriptstyle\downarrow 22.00}$ & 56.67 $\color{red!80!black}{\scriptstyle\downarrow 17.77}$ & 56.52 $\color{red!80!black}{\scriptstyle\downarrow 19.16}$ \\
\midrule

\multirow{6}{*}{5} & \multirow{2}{*}{DeBERTa-v3-xsmall}
& SNLI & 100.00 & 100.00 & 100.00 & 100.00 \\
& & \dataset & 100.00 $\color{black}{\scriptstyle\pm 0.00}$ & 100.00 $\color{black}{\scriptstyle\pm 0.00}$ & 100.00 $\color{black}{\scriptstyle\pm 0.00}$ & 100.00 $\color{black}{\scriptstyle\pm 0.00}$ \\
\cmidrule{2-7}
& \multirow{2}{*}{ELECTRA-small}
& SNLI & 100.00 & 100.00 & 100.00 & 100.00 \\
& & \dataset & 100.00 $\color{black}{\scriptstyle\pm 0.00}$ & 100.00 $\color{black}{\scriptstyle\pm 0.00}$ & 100.00 $\color{black}{\scriptstyle\pm 0.00}$ & 100.00 $\color{black}{\scriptstyle\pm 0.00}$ \\
\cmidrule{2-7}
& \multirow{2}{*}{BERT-base}
& SNLI & 100.00 & 100.00 & 100.00 & 100.00 \\
& & \dataset & 100.00 $\color{black}{\scriptstyle\pm 0.00}$ & 100.00 $\color{black}{\scriptstyle\pm 0.00}$ & 100.00 $\color{black}{\scriptstyle\pm 0.00}$ & 100.00 $\color{black}{\scriptstyle\pm 0.00}$ \\
\bottomrule
\end{tabular}
\caption{Performance comparison after fine-tuning on SNLI versus Atomic-SNLI. Models trained on Atomic-SNLI show substantial improvements for 2- and 3-fact hypotheses, with gains up to 10.07\% accuracy for 3-fact cases. Performance degrades for 4-fact hypotheses due to data sparsity.}
\label{tab:main_results}
\end{table*}

\begin{table*}[t]
\centering
\small
\begin{tabular}{ccccccc}
\toprule
\textbf{Category} & \textbf{Metric} & \textbf{1 Fact} & \textbf{2 Facts} & \textbf{3 Facts} & \textbf{4 Facts} & \textbf{5 Facts} \\
\midrule
\multirow{4}{*}{\textcolor{blue}{Entailment}} 
& Accuracy & 91.23 & 80.59 & 82.55 & 52.00 & 100.00 \\
& Precision & 92.39 & 86.91 & 72.22 & 60.00 & 100.00 \\
& Recall & 90.58 & 64.59 & 44.83 & 25.00 & 100.00 \\
& F1 & 91.48 & 74.11 & 55.32 & 35.29 & 100.00 \\
\midrule
\multirow{4}{*}{\textcolor{orange}{Neutral}} 
& Accuracy & 91.23 & 80.59 & 82.55 & 52.00 & 100.00 \\
& Precision & 87.03 & 73.19 & 82.76 & 50.00 & 0.00 \\
& Recall & 88.07 & 88.80 & 94.74 & 90.00 & 0.00 \\
& F1 & 87.55 & 80.24 & 88.34 & 64.29 & 0.00 \\
\midrule
\multirow{4}{*}{\textcolor{red}{Contradiction}} 
& Accuracy & 91.23 & 80.59 & 82.55 & 52.00 & 100.00 \\
& Precision & 94.00 & 89.79 & 86.36 & 50.00 & 100.00 \\
& Recall & 94.86 & 84.74 & 86.36 & 33.33 & 100.00 \\
& F1 & 94.43 & 87.19 & 86.36 & 40.00 & 100.00 \\
\bottomrule
\end{tabular}
\caption{Class-wise performance of DeBERTa-v3-xsmall fine-tuned on Atomic-SNLI across different numbers of atomic facts. Note the balanced performance across classes compared to simple logical rules (Appendix Table~\ref{tab:simple_rule_detailed}).}
\label{tab:further_atomic_performance}
\end{table*}

\begin{table*}[t]
\centering
\small
\begin{tabular}{p{2cm}p{3cm}p{3cm}p{3cm}p{2cm}}
\toprule
\textbf{SNLI Label}  & \textbf{Premise} & \textbf{Hypothesis} & \textbf{Atomic Fact} & \textbf{Atomic Label} \\
\midrule
\multirow{5}{3cm}{\textcolor{blue}{Entailment}}&\multirow{5}{3cm}{A girl is swinging, rather high, on a swing with blue ropes with lots of trees in the background.} & 
\multirow{5}{3cm}{A girl is swinging, rather high, on a swing with blue ropes.} & \cellcolor[HTML]{E6F0FF}
A girl is swinging. & 
\cellcolor[HTML]{E6F0FF}  Entailment \\
&&&&\\
&& & \cellcolor[HTML]{E6F0FF}
The girl is rather high. & 
\cellcolor[HTML]{E6F0FF}  Entailment \\
&&&&\\
&& & \cellcolor[HTML]{E6F0FF}
The girl is on a swing with blue ropes. & 
\cellcolor[HTML]{E6F0FF}  Entailment \\
\midrule

\multirow{3}{3cm}{\\ \textcolor{orange}{Neutral}}&\multirow{3}{3cm}{\\A group of swimmers jump into a pool.} & 
\multirow{3}{3cm}{\\A group of swimmers jump into a pool during a swim meet.} & \cellcolor[HTML]{E6F0FF}
A group of swimmers jump into a pool. & 
\cellcolor[HTML]{E6F0FF}  Entailment \\
&&&&\\
&& & \cellcolor[HTML]{FFEFE6}
The jump occurs during a swim meet. & 
\cellcolor[HTML]{FFEFE6} Neutral \\
\midrule

\multirow{5}{3cm}{\\ \quad \\\textcolor{red}{Contradiction}}&
\multirow{5}{3cm}{A group of men are sitting and standing a courtyard, some of them are reading books and some are talking.} & 
\multirow{5}{3cm}{\\Men are throwing books and talking outside in a courtyard.} & \cellcolor[HTML]{FFD6CD} 
Men are throwing books. & \cellcolor[HTML]{FFD6CD} 
Contradiction \\
&&&&\\
&& & \cellcolor[HTML]{E6F0FF} 
Men are talking outside. & 
\cellcolor[HTML]{E6F0FF}  Entailment \\  
&&&&\\
&& & \cellcolor[HTML]{E6F0FF} 
Men are in a courtyard. & 
\cellcolor[HTML]{E6F0FF}  Entailment \\
\bottomrule
\end{tabular}
\caption{Atomic-level decomposition examples enabling transparent, explainable reasoning for hypothesis verification.}
\label{tab:atomic_examples}
\end{table*}

To evaluate the effectiveness of Atomic-SNLI, we conduct experiments designed to answer two key questions: (1) Does fine-tuning on atomic-level examples improve a model's ability to perform fine-grained reasoning? (2) Does this atomic-level training transfer to and potentially improve performance on the standard sentence-level NLI task?

Because hypotheses with only one atomic fact do not require fine-grained reasoning, we focus our analysis on instances with hypotheses containing multiple atomic facts.

\subsection{Experimental Setup}

\paragraph{Models.} To eliminate potential biases from other NLI datasets, we do not utilize existing NLI models in this study. Instead, we fine-tune a variety of pre-trained language models on our dataset to evaluate their generalizability. Specifically, we use DeBERTa-v3-xsmall~\cite{he_debertav3:_2021}, ELECTRA-small~\cite{he_debertav3:_2021} and BERT-base~\cite{devlin_bert:_2019}. This selection presents a range of models with different sizes and architectural strengths.

\paragraph{Training Details.} All models are fine-tuned for 3 epochs with a learning rate of $5\times10^{-5}$ and a batch size of 32. The training data for the \dataset~condition is our atomic-level training split. For a fair comparison, baseline models are fine-tuned for the same number of epochs on the original SNLI training set.

\paragraph{Evaluation.} To evaluate the atomic-level training of Atomic-SNLI, we perform experiments on the Atomic-SNLI test set, where the model classifies the relationship between a premise $p$ and each atomic fact $a \in \mathcal{A}_h$. We then sum the predicted probability distributions for all $a \in \mathcal{A}_h$ and take the argmax of the summed probabilities to determine the final label. This provides a less sparse aggregation than the strict logical rules. We compare the accuracy, precision, recall, and F1 score of models trained on Atomic-SNLI and the same models trained on SNLI with the same settings.

\subsection{Main Results}

Table~\ref{tab:main_results} presents our main experimental findings comparing models fine-tuned on standard SNLI versus Atomic-SNLI. The key insight is that atomic-level training significantly enhances performance for multi-fact reasoning:

\paragraph{Consistent Improvements for Multi-fact Hypotheses} Models trained on Atomic-SNLI consistently outperform their SNLI-trained counterparts for 2- and 3-fact hypotheses. DeBERTa-v3-xsmall shows the most dramatic improvement, with +10.07\% accuracy and +8.16\% F1 for 3-fact cases.

\paragraph{Architectural Robustness} The benefits of atomic-level training are observed across different model architectures, with DeBERTa variants showing the largest gains, followed by ELECTRA and BERT.

\paragraph{Data Sparsity Challenges} For 4-fact hypotheses, performance degrades due to limited training examples (only 25 instances in test set), highlighting the need for more complex hypothesis examples in future datasets.

\subsection{Further Analysis}

Our analysis reveals distinct advantages of probability summation over alternative aggregation methods:

\paragraph{Balanced Class Performance} As shown in Table~\ref{tab:further_atomic_performance}, probability summation maintains balanced precision and recall across all three NLI classes. This contrasts with simple logical rules (Appendix Table~\ref{tab:simple_rule_detailed}), which exhibit extreme patterns of high precision/low recall for entailment and low precision/high recall for contradiction.

\paragraph{Error Tolerance Mechanism} The probability summation approach demonstrates robust error handling by:
\begin{itemize}
    \item \textbf{Mitigating Single-Error Catastrophe:} Unlike simple rules where one incorrect atomic prediction can determine the final label, probability summation aggregates confidence scores across all atoms
    \item \textbf{Preserving Uncertainty Information:} Soft probability values retain uncertainty information that is lost in hard logical decisions
    \item \textbf{Enabling Smooth Transitions:} The method provides natural interpolation between clear-cut and borderline cases
\end{itemize}



\subsection{Atomic-Level Explainability}

The atomic decomposition framework is pivotal in enhancing the transparency and interpretability of model reasoning. By dissecting a hypothesis into fundamental atomic facts, our method allows each part of a model's decision to be traced back to specific, individual claims. This level of granularity not only builds trust in the model's conclusions but also effectively identifies the source of any discrepancies, moving beyond the limitations of a ``black box'' approach.

Table~\ref{tab:atomic_examples} demonstrates this process with illustrative examples:

\paragraph{\textcolor{blue}{Entailment} example.}
In the first example, a straightforward entailment relationship is evident. The premise offers all necessary details to support the hypothesis's atomic facts: "A girl is swinging," "The girl is rather high," and "The girl is on a swing with blue ropes," ensuring a clear and verifiable entailment at both atomic and sentence levels.

\paragraph{\textcolor{orange}{Neutral} example.}
The second example captures a neutral scenario. While the atomic fact "A group of swimmers jump into a pool" is entailed by the premise, the additional information "The jump occurs during a swim meet" is not directly supported by the premise, highlighting the neutral nature of this fact.

\paragraph{\textcolor{red}{Contradiction} example.}
In the final contradiction example, the atomic decomposition method efficiently isolates the contradiction. The premise revolves around "men sitting and standing in a courtyard, with some reading books and others talking." This directly opposes the hypothesis that "Men are throwing books," while supporting the facts "Men are talking outside" and "Men are in a courtyard." This finely-tuned analysis provides a definite, fact-based explanation for the contradiction label, advancing beyond the opacity of sentence-level models.


\section{Conclusion}
\label{sec:conclusion}

We presented a comprehensive investigation of atomic-level natural language inference, demonstrating limitations in current models' compositional reasoning capabilities. Our analysis reveals that the conventional assumption of universal atomic entailment fails in practice, with models performing worse on atomic-level inference than sentence-level tasks.

To address this, we introduced \dataset, a novel dataset for fine-grained NLI constructed through decomposition of \snli~and enrichment with carefully generated atomic examples. Experimental results demonstrate that models fine-tuned on \dataset~achieve significant improvements in atomic reasoning while maintaining sentence-level performance.

Our work establishes a new benchmark for granular NLI and provides a path toward more transparent and robust reasoning systems. Future work should explore more sophisticated composition operations and extend atomic-level reasoning to broader textual understanding tasks.

\section*{Limitations}

Our approach has several limitations. First, we rely on LLM-generated contradiction examples, which may not fully capture the complexity of real-world contradictory relationships. Second, our decomposition process depends on the quality of \textsc{DecModel}, which may introduce errors in atomic fact extraction. Third, the current inference rules assume independence between atomic facts, which may not hold in all cases. Future work should address these limitations through more robust generation methods and sophisticated compositional reasoning frameworks.




\bibliography{custom}

\appendix

\section{Detailed Comparison: Simple Rules vs. Sentence-Level Reasoning}
\label{sec:detailed_comparison}

\subsection{Limitations of Simple Logical Rules}

As discussed in Section~\ref{sec:atomic}, simple logical aggregation rules, while theoretically more transparent, demonstrate significant practical limitations. Table~\ref{tab:simple_rule_detailed} shows the detailed results of DeBERTa-v3-xsmall trained on Atomic-SNLI using simple logical rules.

\begin{table*}[t]
\centering
\small
\begin{tabular}{ccccccc}
\toprule
\textbf{Category} & \textbf{Metric} & \textbf{1 Fact} & \textbf{2 Facts} & \textbf{3 Facts} & \textbf{4 Facts} & \textbf{5 Facts} \\
\midrule
\multirow{4}{*}{\textcolor{blue}{Entailment}} 
& Accuracy & 91.23 & 77.07 & 77.85 & 36.00 & 50.00 \\
& Precision & 92.39 & 93.33 & 100.00 & 0.00 & 0.00 \\
& Recall & 90.58 & 49.03 & 20.69 & 0.00 & 0.00 \\
& F1 & 91.48 & 64.29 & 34.29 & 0.00 & 0.00 \\
\midrule
\multirow{4}{*}{\textcolor{orange}{Neutral}} 
& Accuracy & 91.23 & 77.07 & 77.85 & 36.00 & 50.00 \\
& Precision & 87.03 & 70.69 & 81.48 & 35.00 & 0.00 \\
& Recall & 88.07 & 84.27 & 86.84 & 70.00 & 0.00 \\
& F1 & 87.55 & 76.89 & 84.08 & 46.67 & 0.00 \\
\midrule
\multirow{4}{*}{\textcolor{red}{Contradiction}} 
& Accuracy & 91.23 & 77.07 & 77.85 & 36.00 & 50.00 \\
& Precision & 94.00 & 79.26 & 70.97 & 40.00 & 100.00 \\
& Recall & 94.86 & 95.18 & 100.00 & 66.67 & 100.00 \\
& F1 & 94.43 & 86.50 & 83.02 & 50.00 & 100.00 \\
\bottomrule
\end{tabular}
\caption{Detailed performance analysis using simple logical aggregation rules. Note the high-precision low-recall pattern for entailment and low-precision high-recall pattern for contradiction, specifically for 3-fact hypotheses.}
\label{tab:simple_rule_detailed}
\end{table*}

From the table, we observe the following key patterns:

\paragraph{Entailment Class Imbalance} Under simple rules, entailment exhibits extremely high precision but very low recall. This occurs because the rule requires \textbf{all} atomic facts to be entailed—any single atomic error causes the entire hypothesis to be rejected. In the 3-fact case, precision reaches 100\% while recall is only 20.69\%, indicating the rule is overly strict.

\paragraph{Contradiction Class Sensitivity} The contradiction class shows high recall but relatively low precision because the rule states that \textbf{any} atomic fact contradiction leads to the entire hypothesis being labeled as contradiction. This "one-vote veto" mechanism makes the model highly sensitive to noise and errors.

\paragraph{Neutral Class Complexity} The neutral class has the most complex definition under simple rules, requiring "no contradiction and at least one neutral fact," leading to performance instability across different atomic fact counts.

\subsection{Advantage Mechanisms of Probability Summation}

The probability summation rule alleviates the strictness limitations of simple rules through soft aggregation. Its core advantages include:

\begin{itemize}
    \item \textbf{Error Tolerance:} A single atomic prediction error does not directly determine the final label
    \item \textbf{Confidence Integration:} Incorporates confidence information from all atomic predictions
    \item \textbf{Smooth Decision Boundaries:} Provides smoother decision boundaries in borderline cases
\end{itemize}

\subsection{Comparison with Direct Sentence-Level Reasoning}

\begin{table*}[t]
\centering
\small
\begin{tabular}{ccccccc}
\toprule
\textbf{Category} & \textbf{Metric} & \textbf{1 Fact} & \textbf{2 Facts} & \textbf{3 Facts} & \textbf{4 Facts} & \textbf{5 Facts} \\
\midrule
\multirow{4}{*}{\textcolor{blue}{Entailment}} 
& Accuracy & 91.13 & 88.42 & 89.26 & 72.00 & 100.00 \\
& Precision & 92.48 & 92.41 & 90.91 & 100.00 & 100.00 \\
& Recall & 90.16 & 80.54 & 68.97 & 66.67 & 100.00 \\
& F1 & 91.31 & 86.07 & 78.43 & 80.00 & 100.00 \\
\midrule
\multirow{4}{*}{\textcolor{orange}{Neutral}} 
& Accuracy & 91.13 & 88.42 & 89.26 & 72.00 & 100.00 \\
& Precision & 86.66 & 84.24 & 88.89 & 61.54 & 0.00 \\
& Recall & 88.11 & 91.20 & 94.74 & 80.00 & 0.00 \\
& F1 & 87.38 & 87.58 & 91.72 & 69.57 & 0.00 \\
\midrule
\multirow{4}{*}{\textcolor{red}{Contradiction}} 
& Accuracy & 91.13 & 88.42 & 89.26 & 72.00 & 100.00 \\
& Precision & 93.98 & 91.63 & 89.13 & 50.00 & 100.00 \\
& Recall & 94.97 & 92.37 & 93.18 & 66.67 & 100.00 \\
& F1 & 94.47 & 92.00 & 91.11 & 57.14 & 100.00 \\
\bottomrule
\end{tabular}
\caption{Detailed performance analysis using sentence-level inference.}
\label{tab:sentence_detailed_comparison}
\end{table*}

Table~\ref{tab:sentence_detailed_comparison} presents a detailed performance of direct sentence-level reasoning. We can compare it with and our atomic-level approach in Table~\ref{tab:further_atomic_performance}. Key findings include:


\paragraph{Trade-off in Simple Cases} For 2- and 3-fact cases, both methods show comparable performance, but the atomic-level approach offers better interpretability. For 4-fact cases, both methods face challenges due to data sparsity.

\paragraph{Value of Interpretability} Although performance is similar in some simple cases, the core advantage of the atomic-level method lies in the transparent decision process it provides, which is crucial for trust and debugging in practical applications.

\section{Contradiction Generation Prompt}
\label{sec:prompt}

To construct high-quality contradiction pairs for our Atomic-SNLI dataset, we designed a specialized prompt in Figure \ref{fig:contradiction_prompt} for a large language model. The objective is to generate a minimally altered version of an atomic fact that directly contradicts a given premise.

\begin{figure*}[t]
\materialCards{Contradiction Generation Prompt}{
You are a linguistic contradiction generator. Your task is to create a minimally modified version of a given atomic fact that contradicts the premise, while maintaining grammaticality and semantic coherence.

\textbf{Instructions:}
\begin{enumerate}
    \item Create a contradiction by modifying \textbf{ONLY 1--2 key elements} in the atomic fact.
    \item Maintain the same grammatical structure and length.
    \item Ensure the modified fact directly contradicts the premise.
    \item Keep high lexical similarity with the original fact.
    \item The contradiction must be clear and unambiguous.
\end{enumerate}

\textbf{Example 1:}\\
\medskip
\textbf{Premise:} A soccer game with multiple males playing.\\
\textbf{Original Atomic Fact:} Some men are playing a sport.\\
\textbf{Contradiction:} No men are playing a sport.\\

\medskip
\textbf{Example 2:}\\
\medskip
\textbf{Premise:} A woman is playing the guitar in a park.\\
\textbf{Original Atomic Fact:} A person is playing an instrument.\\
\textbf{Contradiction:} No person is playing an instrument.\\

\medskip
\textbf{Task:}\\
\textbf{Premise:} \{premise\}\\
\textbf{Original Atomic Fact:} \{atomic\_fact\}\\
\textbf{Contradiction:}\\
}
\caption{The prompt used to generate contradiction pairs for atomic facts. The model is instructed to produce a minimal edit that creates a direct and unambiguous contradiction with the premise.}
\label{fig:contradiction_prompt}
\end{figure*}

\end{document}